# Model Retirement Creates Reproducibility Risk in Biomedical AI Publications

Nathan Wolfrath, MD[1], Meghan Conroy, MD[1], Thomas Kosten, BS[1], Dave Bell, BA; Bhabishya Neupane, BS; Jonah Kindel, Anjishnu Banerjee, PhD[2], Priya Deshpande, PhD[3], Bradley Taylor, MBA[4], and Anai N. Kothari, MD, MS[1]

[1]*Bud and Sue Selig Hub for Surgical Data Science, Department of Surgery, Medical College of Wisconsin*
[2]*Division of Biostatistics, Data Science Institute, Medical College of Wisconsin*
[3]*Department of Computer and Electrical Engineering, Marquette University*
[4]*Clinical and Translational Science Institute, Medical College of Wisconsin, Milwaukee*

Corresponding Author:

Anai N Kothari, MD, MS
Department of Surgery/ Division of Surgical Oncology
8701 Watertown Plank Road
Milwaukee, WI 53226
Phone: 414-955-1443
FAX: 414-955-0197
Email: akothari@mcw.edu

## ABSTRACT

**Background**. Large language models (LLMs) are being adopted in biomedical research at a rapid and accelerating pace, yet commercial services that host many widely used models operate under deprecation schedules that can complicate scientific reproducibility.

**Methods**. We searched PubMed for original research articles from 2022 through March 2026 that applied a specific LLM to a biomedical task. An extraction agent identified model names from 61,077 article abstracts with human reviewers validating a subset for extraction accuracy. Extracted model names were normalized to canonical model identifiers. Lifecycle data (release date, retirement date, status) were compiled for the 50 most frequently used models.

**Results**. We identified 8,931 paper-model mentions spanning 5,242 unique publications after restricting the analysis to the 50 most frequently used models. Among these mentions, 77.7% cited a commercial closed-weight model. Overall, 42% involved a model that was already retired by the time of official publication or is scheduled to retire within two years of publication. The median interval from publication to model retirement was 538 days.

**Conclusion**. Many biomedical publications using LLMs are on a trajectory toward computational non-reproducibility after publication. Model deprecation should be treated as a core reporting and preservation issue for biomedical research.

## INTRODUCTION

Large language models (LLMs) have been applied across a diverse range of biomedical research including clinical decision support, diagnostics, patient interaction, clinical documentation, drug discovery, and medical education.[1], [2], [3], [4] Furthermore, the medical LLM literature has been rapidly expanding.[5] A recent review identified over 4,600 peer-reviewed studies evaluating LLMs in clinical medicine between January 2022 and September 2025, with ChatGPT and related OpenAI models constituting approximately 66% of evaluated models.[6] Traditional biomedical machine-learning workflows often rely on versioned statistical software, open-source libraries, and locally executable code. In contrast, many contemporary LLM workflows depend on commercial products accessed through user interfaces or application programming interfaces (APIs).[7], [8] In either of these consumption methods, researchers often have limited control over updates, behavioral changes, or deprecations affecting the underlying model.

Reproducibility is a cornerstone of evidence-based medicine and the scientific process more broadly. Recent surveys of the international biomedical research community indicate that there is a reproducibility crisis in medicine.[9] The inability to re-execute a published workflow reduces reliability, limits external validation, and impedes extension in future work. Reproduction of LLM-related work often depends heavily on precise reporting of exact model names, version, access date, and prompts.[10] However, even with proper reporting of these features, rapid release of new models and deprecation of prior generations raises concern for a reproducibility gap unique to model-dependent research.[11]

In this work, we present a systematic bibliometric characterization of this reproducibility problem. We quantify how many LLM-citing biomedical papers reference models that are already retired or will retire soon after publication.

## METHODS

### *Literature Search*

We queried PubMed (NCBI E-utilities API) for biomedical research articles published between January 2022 and July 2026 containing the terms *large language model, LLM, generative AI, generative artificial intelligence, artificial intelligence, foundation model,* or *language model* in the title or abstract. Reviews, systematic reviews, meta-analyses, editorials, letters, and comments were excluded via PubMed publication-type filters. Works which did not specify publication date beyond the year of publication were excluded (Figure 1).

### *Model Mention Extraction*

Model extraction was performed using a GPT-4o-based agent (GPT-4o-2024-08-06) via Azure OpenAI with temperature = 0 to determine whether each abstract described the applied use of an existing named LLM in a biomedical task. Papers whose primary contribution was a novel model or architecture were excluded. This agent was prompted to return either the most specific version strings available or an indication of no qualifying use. If multiple models were used in a single publication, each paper-model mention was considered a separate unit in this analysis. A subset of 50 processed papers were reviewed by humans to verify extraction accuracy.

*Name Normalization*

Extracted raw model strings were processed through a normalization pipeline to harmonize spelling variants and other references to the same model. Encoder-only models (e.g., BERT-family models) and protein language models (e.g., ESM-family models) were excluded at this step in analysis. Generic model family names without version information (e.g., "Claude" or "ChatGPT") were excluded as they were insufficiently specific to map to a single model. This choice follows broader reporting consensus that exact model identification rather than family-level labels are required.[12], [13]

*Lifecycle Data*

Release dates, retirement dates, and current status were retrieved for the 50 most mentioned models were retrieved from primary vendor sources (eg OpenAI deprecation pages [14], Anthropic model deprecation logs [15]). Status was determined to be active (currently available with no announced retirement date), scheduled (retirement announced but future-dated), retired, or open weight (weights publicly available; no formal retirement). Models were classified by license type as commercial, open weight, or untracked-commercial products (consumer products such as OpenEvidence or Microsoft Copilot with no per-version release or retirement schedule).

### *Analysis*

For each paper-model mention, we computed the number of days between publication date and underlying model retirement date. Open weight models are treated as having no retirement date. Publication dates were collapsed to the month of publication for time series analysis. We use a cutoff of model retirement within two years of publication date to define a subset of papers which are rapidly outdated shortly after publication.

## RESULTS

### *Corpus Characteristics*

PubMed search yielded 65,955 successful abstract retrievals. Of these, 8,506 (13.9%) contained at least one qualifying model mention in the abstract, yielding 16,216 raw mention strings. After normalization and exclusion of encoder-only models, generic family names, and non-model tool references, 8,931 mentions from 5,242 papers remained. Human review by two reviewers of a subset of 50 extractions gave an average extraction accuracy of 96.2%. Of extraction errors, the majority (83.3%) were missed extractions rather than hallucnations of models not actually used. Restricting to the top 50 models with compiled lifecycle data produced the primary analytic dataset: 8,931 paper-model mentions across 5,242 unique publications (Figure 1).

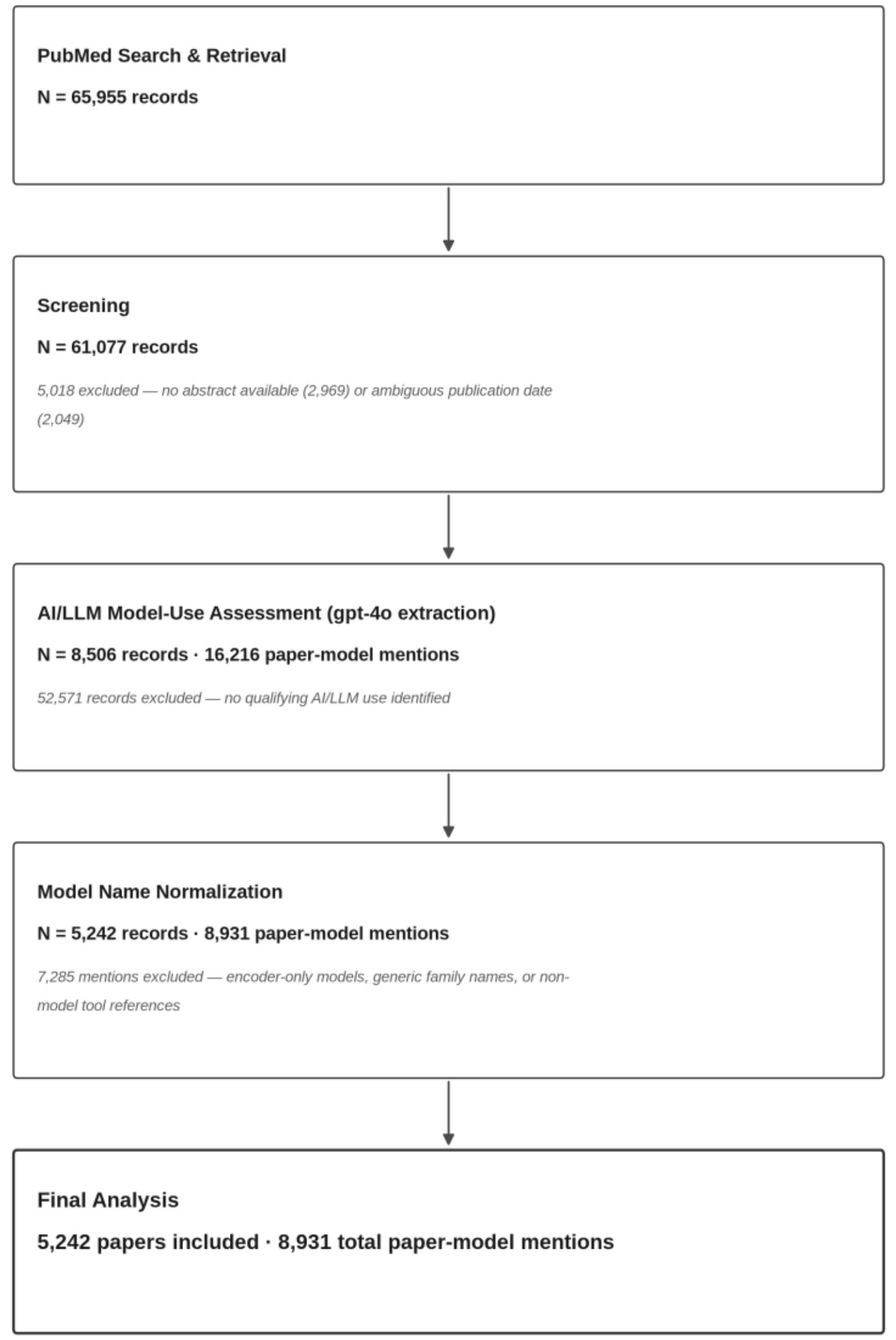

Mention volume increased steeply: 9 in 2022, 399 in 2023, 2,083 in 2024, 4,570 in 2025, and 1,870 in the first quarter of 2026 (Figure 2). This trajectory is consistent with prior bibliometric analysis documenting exponential growth in LLM-related biomedical publications over the same period.[16], [17]

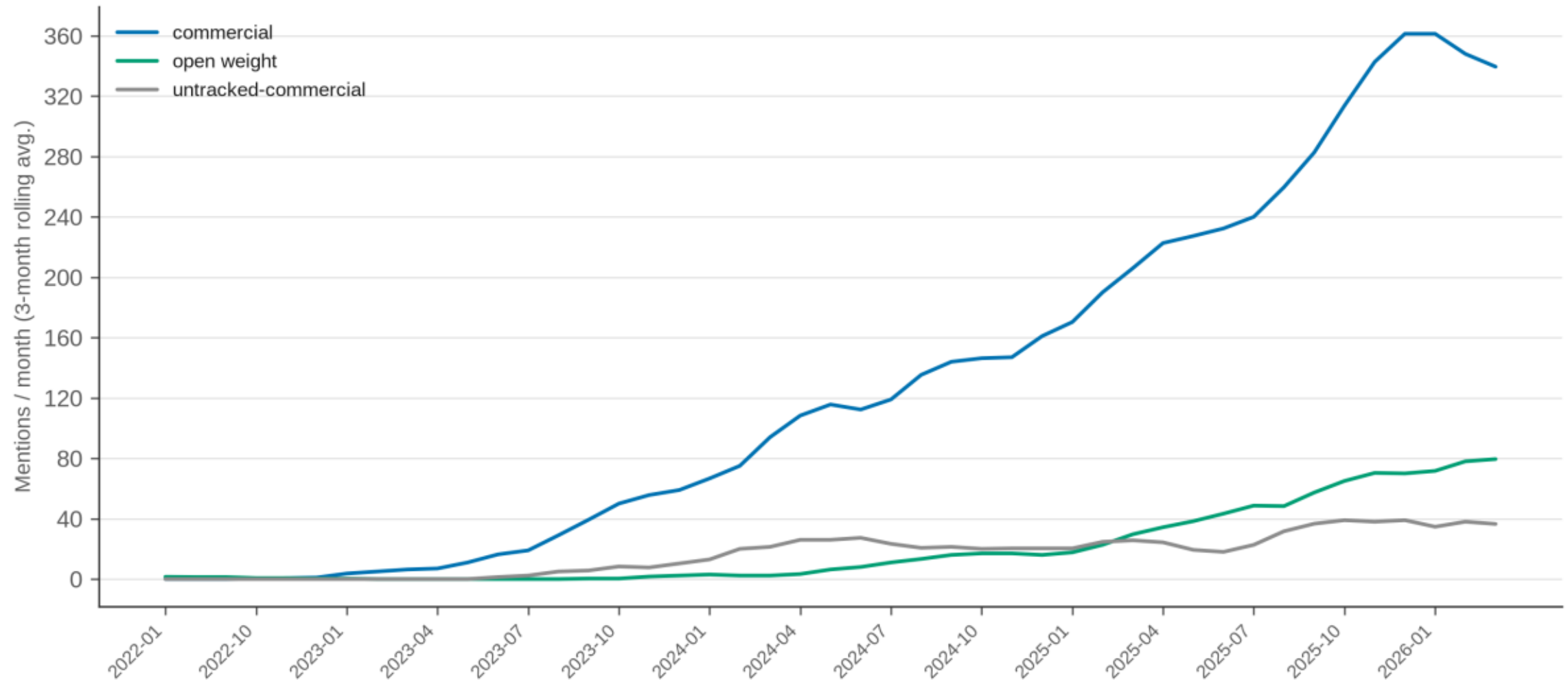

*License and Vendor Distribution*

Commercial API models accounted for 6,936 mentions (77.7%), open-weight models for 1,089 (12.2%), and untracked-commercial products for 906 (10.1%, Figure 2). OpenAI models dominated across all categories: 5,864 mentions (65.7% of in-scope total). The three most-cited models were GPT-4 (2,480 mentions; 27.8%), GPT-4o (1,318; 14.8%), and GPT-3.5-turbo (1,266; 14.2%) (Figure 3); all commercial APIs with formally announced retirement dates.

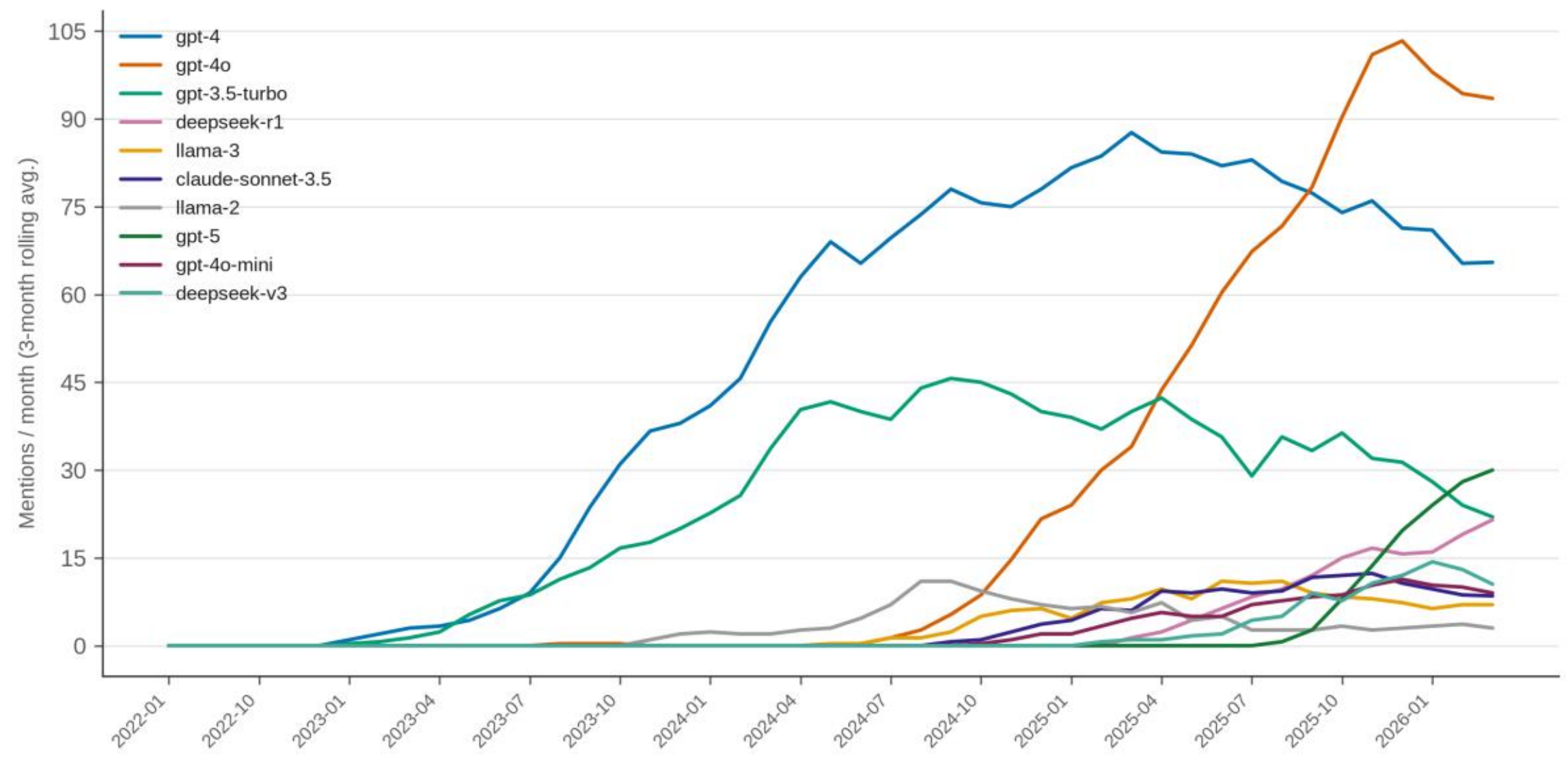

*Reproducibility Risk Due to Model Retirement*

Among all 8,931 in-scope mentions, 3,715 (41.6%) involved a model with a confirmed retirement date falling within 730 days of the citing paper's publication date. At publication, 256 mentions (2.9%) cited a model that had already been retired. Among the 898 mentions with both a confirmed retirement date and retired status, 256 (28.5%) were published after the model was withdrawn. The mean interval between publication and retirement across all retired and scheduled mentions was 546 days (approximately 17.9 months; Table 1).

Two cases illustrate the pattern. Claude Sonnet 3.5, released in June 2024 and retired in October 2025, had 33.1% of its 172 citing papers appear after retirement. Base version GPT-5, released in August 2025 and scheduled for retirement in December 2026, is already cited in 182 papers, all of which will meet our definition of reproducibility risk due to model retirement upon its sunset.

## DISCUSSION

This analysis reveals a mismatch between the temporal assumptions embedded in biomedical publishing and the operational longevity of commercial LLM services. Peer review and publishing process alone may consume a substantial portion of a commercial model's supported lifespan. By the time a paper is indexed, the workflow it describes may already be difficult or impossible to replicate. More broadly, commercial LLM studies face reproducibility problems tied to deprecated model access, incomplete artifacts, and evolving provider ecosystems.[18], [19]

This reproducibility risk is amplified by the predominance of commercial models. The appeal of established commercial vendors is understandable, including cutting-edge model performance and mature tooling (and indeed, an OpenAI model was used in this study). However, open-weight models carry a significant advantage of persistent and accurate model versioning.[20] Additionally, while a large array of benchmarks exist, open-weight models have achieved performance competitive with proprietary systems in some specific tasks such as medical evidence summarization, clinical decision-making, and radiology report extraction.[21], [22], [23] Commercial APIs generally do not provide the same preservation guarantee, yet they constituted 77.7% of LLM citations in our corpus. The dominance of OpenAI models (65.7% of in-scope mentions) means that a single commercial vendor's retirement schedule contributes substantially to reproducibility risk across the biomedical literature.

Several interventions follow from these findings. First, journals should require authors to report a specific version string (e.g., gpt-4o-2024-08-06, not merely "GPT-4o"), just as clinical studies must report drug lot numbers and assay kit versions. Reporting checklists for AI-based biomedical studies have begun to incorporate such requirements, but adherence remains inconsistent.[13], [24] Second, for tasks requiring long-term reproducibility such as clinical guideline support or regulatory submissions, open-weight models should be preferred, and weights should be archived alongside the manuscript in a persistent repository. Third, preprint and other publishers could surface a structured field for model version and query date, enabling easier tracking of the reproducibility lifecycle of biomedical AI literature.

Several limitations must be noted. First, our analysis was restricted to abstract-level extraction. A more specific model identifier may have appeared in the main text or supplement of each work. Second, LLM-based extraction introduces the possibility of error. Although a language model is well suited to extract model names from short abstracts, misclassifications may occur.[25] Human verification, however, did reveal high-accuracy extraction. Third, although lifecycle dates were compiled from primary vendor sources where available, retirement schedules are subject to change.

Additionally, assuming that models are static is operationally conservative but may understate the scope of the problem. In practice, commercial models that nominally remain active may have their underlying weights updated silently, their API endpoints redirected, or their system-level behavior altered through undisclosed instruction tuning or alignment updates.[11], [26] These behavior changes are invisible to our analysis. Some popular products are inherently non-versioned and provide no visible tracking of changes to the underlying models, which makes longitudinal tracking difficult even when a product name is reported.[12] Beyond the same model weights, true reproducibility of a study additionally requires full execution context, including parameters such as temperature, prompt, and output schema.[27]

The pace of model releases and use in biomedical literature is accelerating. Without careful action from publishers, researchers, and vendors, a growing share of the biomedical AI literature will describe experiments that cannot be verified, audited, or built upon, eroding the foundation needed for confident and safe application.

## DATA AVAILABILITY

Code can be made available upon reasonable request to the corresponding author. The PubMed abstracts used in this work are publicly available.

## CONFLICT OF INTEREST

The authors declare no conflicts of interest.

## AI DISCLOSURE

During preparation of this manuscript, the authors used ChatGPT (OpenAI) and Claude (Anthropic) to assist with language refinement, organization, and editing of the manuscript. All AI-assisted content was reviewed and edited by the authors, who take full responsibility for the accuracy, integrity, and final content of the manuscript.

## TABLES

**Table 1. Usage and retirement status of large language models reported in the published literature.** Models are ranked by the number of identified publications in which they were used. Retirement status and announced retirement dates are shown for commercial models; open-weight models without a defined vendor retirement date are designated separately. "Avg. days to retire" represents the mean interval between publication and model retirement, with negative values indicating publication after retirement. "% <2 yr to retire" represents the proportion of publications occurring within two years of the model's retirement date. "Retired at publication" indicates the number of papers published after the referenced model had already been retired. Commercial products without a single trackable model version were excluded.

| Model | Vendor | License | Status | Retirement date | Papers | Avg. days to retire | % <2yr to retire | Retired at publication |
|---|---|---|---|---|---|---|---|---|
| **gpt-4** | OpenAI | commercial | Scheduled | 2026-10-23 | 2,466 | +660d | 62% | 0 |
| **gpt-4o** | OpenAI | commercial | Active (no date) | — | 1,302 | — | — | — |
| **gpt-3.5-turbo** | OpenAI | commercial | Scheduled | 2026-10-23 | 1,256 | +708d | 54% | 0 |
| **deepseek-r1** | DeepSeek | open weight | Open weight | — | 193 | — | — | — |
| **claude-sonnet-3.5** | Anthropic | commercial | Retired | 2025-10-28 | 172 | +98d | 67% | 57 |
| **gpt-5** | OpenAI | commercial | Scheduled | 2026-12-11 | 170 | +345d | 100% | 0 |
| **llama-3** | Meta | open weight | Open weight | — | 153 | — | — | — |
| **llama-2** | Meta | open weight | Open weight | — | 133 | — | — | — |
| **gpt-4o-mini** | OpenAI | commercial | Active (no date) | — | 124 | — | — | — |
| **deepseek-v3** | DeepSeek | open weight | Open weight | — | 115 | — | — | — |
| **llama-3.1** | Meta | open weight | Open weight | — | 107 | — | — | — |
| **gpt-4-turbo** | OpenAI | commercial | Scheduled | 2026-10-23 | 102 | +551d | 78% | 0 |
| **gemini-1.5-pro** | Google | commercial | Retired | 2025-09-24 | 101 | +58d | 57% | 43 |
| **o1** | OpenAI | commercial | Scheduled | 2026-10-23 | 96 | +401d | 100% | 0 |
| **gemini-2.0** | Google | commercial | Active (no date) | — | 85 | — | — | — |
| **gemini-2.5-pro** | Google | commercial | Scheduled | 2026-10-20 | 83 | +322d | 100% | 0 |
| **gemini-advanced** | Google | commercial | Active (no date) | — | 77 | — | — | — |
| **claude-opus-3** | Anthropic | commercial | Retired | 2026-01-05 | 70 | +255d | 84% | 9 |
| **grok-3** | xAI | commercial | Retired | 2026-05-15 | 69 | +184d | 100% | 0 |
| **claude-2** | Anthropic | commercial | Retired | 2025-07-21 | 67 | +213d | 81% | 13 |
| **gemini-2.0-flash** | Google | commercial | Retired | 2026-06-01 | 65 | +218d | 100% | 0 |
| **gpt-3** | OpenAI | commercial | Retired | 2024-01-04 | 63 | -37d | 48% | 32 |
| **claude-sonnet-3.7** | Anthropic | commercial | Retired | 2026-02-19 | 61 | +81d | 79% | 13 |
| **qwen-2.5** | Alibaba | open weight | Open weight | — | 55 | — | — | — |
| **llama-3.3** | Meta | open weight | Open weight | — | 54 | — | — | — |

## FIGURE LEGENDS

**Figure 1. Study identification, screening, and model-use assessment workflow.** Flow diagram depicting identification and screening of PubMed-indexed records, AI/LLM model-use assessment, model name normalization, and inclusion in the final analysis.

**Figure 2. Model mentions by license category over time.** Monthly mentions of artificial intelligence models in the identified literature are shown according to model license category: commercial, open weight, and untracked-commercial. Values represent 3-month rolling averages of model mentions per month.

**Figure 3. Trends in use of the 10 most frequently mentioned trackable models in PubMed-indexed biomedical research.** Monthly mentions of the 10 most frequently identified trackable AI/LLM models are shown over time. Values represent 3-month rolling averages of model mentions per month. Models are ranked based on their total number of mentions across the study period.